\documentclass{ifacconf}

\usepackage{graphicx}      
\usepackage{natbib}        
\usepackage{amsmath}
\usepackage{amssymb}
\usepackage{algpseudocode}
\usepackage{algorithm}
\usepackage{dsfont}
\usepackage{textcomp}
\usepackage{makecell}
\usepackage{multicol}
\usepackage{xurl}
\usepackage{subcaption}
\usepackage[utf8]{inputenc}
\usepackage[T1]{fontenc}
\usepackage{xcolor}

\algrenewcommand\algorithmicindent{0em}
\algrenewcommand\algorithmicrequire{\textbf{Input:}}
\algrenewcommand\algorithmicensure{\textbf{Output:}}
\newcommand{\algorithmicbreak}{\textbf{break}}
\newcommand{\Break}{ \algorithmicbreak}
\def\GR#1{{#1}}
\def\JP#1{{#1}}
\def\JPL#1{{#1}}

\begin{document}

\setlength\abovedisplayskip{8pt}%
\setlength\belowdisplayskip{8pt}%
\setlength\abovedisplayshortskip{8pt}%
\setlength\belowdisplayshortskip{8pt}%

\begin{frontmatter}

\title{A cell-decomposition based path planner for 3D navigation in constrained workspaces\thanksref{footnoteinfo}} 

\thanks[footnoteinfo]{This manuscript version is made available under the CC-BY-NC-ND 4.0 license. This work was supported by the Brazilian agencies CAPES through the Academic Excellence Program (PROEX), CNPq under the grants 317058/2023-1 and 422143/2023-5, and FAPEMIG. Marcelo A. Santos acknowledges support from the Lombardy Region under the PR FESR 2021–2027 "Collabora \& Innova" call (Decree no. 11969, 2 August 2024), Project "HARMONY" (CUP: E59I25000850007).}

\author[First]{Jo\~{a}o P. L. Morais} 
\author[First,Third]{Luciano C. A. Pimenta}
\author[Second]{Marcelo A. Santos} 
\author[First,Third]{Guilherme V. Raffo}

\address[First]{Graduate Program in Electrical Engineering, Universidade Federal de Minas Gerais, Belo Horizonte, MG, Brazil}
\address[Second]{Department of Management, Information and Production Engineering, University of Bergamo, Dalmine, BG, Italy}
\address[Third]{Department of Electronic Engineering, Universidade Federal de Minas Gerais, Belo Horizonte, MG, Brazil}


\begin{abstract}                
This \JP{paper} \GR{proposes} a cell decomposition algorithm for binary occupancy grids that ensures mutual complete visibility \JP{from each cell to at least one adjacent cell\GR{. This decomposition establishes}} a simplified framework \GR{for verifying} path feasibility \JP{that} can be \JP{easily} embedded in optimization problems. To illustrate \GR{its utility, we formulate both} 
\JP{second-order cone programs (SOCP) and their mixed-integer variant (MISOCP) within the proposed framework. Furthermore, \GR{we propose the KSP-SOCP method, which} combines Yen's k-shortest path algorithm \JP{with} the SOCP\GR{, achieving} improved solutions compared to a standard SOCP approach while avoiding the computational burden of MISOCP}. \JP{The cell decomposition algorithm, KSP-SOCP\GR{,} and MISOCP approaches \GR{were} evaluated in 9 city-like workspaces. The decomposition efficiently partitioned each map\GR{, enabling} both optimization methods to compute feasible paths. The proposed KSP-SOCP achieved time performance comparable to the MISOCP while requiring less memory, making it \GR{highly} suitable for large-scale problems}. 


\end{abstract}


\end{frontmatter}

\section{Introduction}

Path planning is a core challenge for autonomous systems, playing a key role in their navigation in dynamic workspaces and under sensing uncertainty and online planning constraints \citep{Toit2012, Katracazas2015}. Path planners typically use high-level map representations, either unstructured, as in sampling-based methods like RRT \citep{Karaman2011}, or structured, as in cells or roadmap-based approaches \JP{for deterministic path generation \citep{Choset2005,Lupascu2019}}.

\JP{Within the structured path planning methods, a frequent approach is to employ mixed-integer programming (MIP) to either model collision avoidance constraints \citep{Schouwenaars2001,Mellinger2012} or to select a sequence of convex safe sets that the path is to remain within \citep{Deits2015,Marcucci2023}. Such MIPs can ensure optimality, but become intractable for large-scale problems without convex relaxation \citep{Marcucci2024MISO}. Alternatively, heuristic approaches employ line graphs to identify a feasible low-cost sequence of convex safe regions. The resulting path, computed via convex optimization, trades optimality for reduced computational complexity \citep{Marcucci2024}.}

\JP{Recent methods to generate convex safe sets for the aforementioned path planning approaches include the iterative region inflation by semidefinite programming (IRIS) \citep{Deits2015IRIS} and the safe corridor generation based on voxel grids \citep{Toumieh2022}. The former method requires semi-algebraic representations of obstacle sets, while the latter requires an occupancy grid. In both cases, a set of intersecting convex safe regions is obtained, with such intersections being used as nodes in their respective line graphs.}

\JP{Despite their contributions, the aforementioned works still present significant gaps regarding path planning within cell-decomposed workspaces. In the convex safe set generation context, the presented approaches rely on intersections between convex sets to obtain safe regions for path optimization, which excludes scenarios where non-overlapping cell decompositions are more suitable, such as coverage paths. Although \citet{Nielsen2019} \GR{have presented} a non-overlapping cell decomposition for 2D polygonal workspaces where the union of each adjacent cell pair is convex, allowing the generation of safe paths in the same fashion, no works providing a similar result for 3D or higher dimension workspaces were found.} 

\JP{In the path optimization context, heuristic approaches are constrained to the initially selected cell sequence found by the graph traverse algorithm, while MIPs become intractable in large-scale scenarios. However, optimization approaches that extend the heuristic approach by searching and testing additional cell sequences could achieve the optimality guarantees of MIPs without its computational complexity. The optimization approaches are also formulated for specific objectives, such as shortest length or minimum-fuel path \citep{Schouwenaars2001}, without providing a general framework for path optimization that can accommodate different objectives.}

\JP{\GR{To address these gaps, this paper generalizes our} path planning method \GR{presented} in \citet{Morais2025}. In \GR{the new} method, \GR{we propose} a cell decomposition algorithm for 3D binary occupancy grids that generates non-overlapping\GR{,} axis-aligned\GR{,} box-shaped cells \GR{subject to} boundary uniformity constraints. \GR{This design} choice\GR{, which is supported by the performance and practicality arguments} presented in \citet{Marcucci2024} and \citet{Toumieh2022}\GR{, respectively, ensures complete} pairwise visibility \GR{between adjacent cells. This visibility }property \GR{yields} results analogous to \GR{those found in} \citet{Nielsen2019} for 3D environments. \GR{Furthermore, the} extended method \GR{solves certain} issues related to the preservation of Euclidean space connectivity observed in the original cell decomposition procedure.}

\JP{This paper also introduces simplified path feasibility constraints applicable to a wide range of optimization problems, \GR{thereby creating} a general framework for optimization-based path planning. \GR{We demonstrate this} framework \GR{by formulating} second-order cone programs (SOCP) and their mixed-integer variant (MISOCP) for shortest path problems. Furthermore, \GR{we propose the KSP-SOCP method, which} combines Yen's k-shortest path algorithm with an SOCP solver to explore multiple feasible cell sequences\GR{. This approach improves} solution quality while mitigating the computational burden \GR{associated with} the MISOCP formulation.} 

\JP{For comparison purposes, both \GR{the} KSP-SOCP and MISOCP approaches are tested against a Basic Theta* path planner \citep{Kenny2010}\GR{. Theta*} is an any-angle path planner for grids that provides near-optimal shortest paths with time performance comparable to A* \citep{Hart1968}. \GR{This comparison provides} an estimate for the optimality gap of the paths found by KSP-SOCP and MISOCP, and \GR{highlights the difference in} performance between the proposed path planning framework and a \GR{standard} grid-based search on the occupancy grid.}

\JP{The remainder of this paper is organized as follows. Section 2 presents the extended cell decomposition algorithm. Section 3 introduces the path-feasibility framework and the SOCP, MISOCP, and KSP-SOCP formulations. Section 4 presents simulation results in city-like workspaces, and Section 5 concludes the paper.}

\section{Cell decomposition} \label{sec:cellDecomposition}
\JP{Consider} a 3D Euclidean space \( Y \subset \mathbb{R}^{3} \), containing an unknown number \( N_{o} \) of axis-aligned, box-shaped \JP{obstacle} regions \( O_{i} \). Let \( O \) be the union of all \JP{obstacles}, and \( Y_{f} \) \GR{be} the \JP{free} space within \(Y\).
\GR{The space} \( Y \) is discretized \GR{into} a binary occupancy grid \( D \in \{0,1\}^{N_{x} \times N_{y} \times N_{z}} \), where \( D_{ijk} = 1\) if it intersects \JP{an} obstacle, and 0 otherwise. \( D \) is \GR{then} decomposed into \( N_c \) axis-aligned box-shaped cells\GR{,} represented by a matrix \( C \in  \mathbb{N}^{6\times N_{c}}  \) and a Boolean vector \(o \in \{0,1\}^{N_{c}}\). For the $n$-th cell, the first three elements in column \(n\) are the voxel indices of its lowermost corner, and the last three are its uppermost corner. \JP{Furthermore}, \JP{\(o_n\) indicates whether the cell is an obstacle (\(o_{n} = 1\)) or not}.

All cells in $C$ must satisfy the following properties:
\begin{enumerate}
    \item[P1)] \textit{No cell overlap}: no intersection between the internal volumes of any cells;\label{property1}
    \item[P2)] \textit{Internal uniformity}: all voxels inside a cell must be uniformly 1 or 0, ensuring an exact decomposition relative to \( D \);\label{property2}
    \item[P3)] \textit{Mutual complete visibility}: All cells must have mutual complete visibility with at least one adjacent cell of the same type (unless there are \JP{none}). \JP{In this work, two cells \(C_i, C_j\subset Y_f \) are said to be in complete visibility if, for every pair of points \(a \in C_i\) and \(b \in C_j\), the line segment joining \(a\) and \(b\) lies entirely within the free space \(Y_f\); } \label{property3}
    \item[P4)] \textit{Face boundary uniformity}: \JP{all cells' boundaries must be uniform to prevent partial obstructions, i.e., all voxels adjacent to that boundary share the same occupancy value.}\label{property4}
\end{enumerate}
\JP{P1 implies that \GR{every} voxel in \(D\) is covered by a single cell and that cell expansion occurs only \GR{into} uncovered regions of \(D\). P2 implies that all free cells are safe and that \GR{the entire} free space in \(D\) is included \GR{within} a free cell. P3 implies that it is possible to build a roadmap \GR{connecting} all reachable free cells by placing a representative point in each cell and connecting them based on whether their cells satisfy P3. \GR{This property also means} that each representative point can be adjusted to match any given start and goal point within the reachable cells, \GR{thereby enabling} both path planning and optimization through the roadmap. P4 implies that there is at most one portal per cell boundary, and if a portal exists, it spans the entire face boundary. \JP{\GR{Furthermore}, it} implies that for any cell \GR{satisfaying} P1-P4, it is possible to create a unit volume cell on each uniform neighborhood that \GR{also} satisfies P1-P4, which is crucial for the proper execution of the proposed cell decomposition.} \JPL{With all cells satisfying P1-P4, the resulting cell decomposition can ensure resolution-limited topology preservation of \(Y\), since it entirely preserves the topology of the occupancy grid \(D\), and thus its connectivity is only limited by the latter's resolution. }

The major change to the algorithm \JP{proposed} in \citep{Morais2025} is the integration of property P3 directly into the decomposition process. Previously, P3 was only checked after decomposition, which could \GR{result in} a feasible cell \GR{being} disconnected from the graph even if it was physically reachable. The extended algorithm now enforces P3 during cell formation to prevent this issue.

The proposed cell decomposition procedure\footnote{For a visual demonstration of the cell decomposition procedure, please refer to the following video: \url{https://youtu.be/-yTLp6bC9_o}.} \GR{used} to generate $C$ is outlined in Algorithm \ref{alg:cellDecomposition}, and utilizes the following auxiliary variables: 
\begin{itemize}
    \item \( B \in \mathbb{N}^{N_{x} \times N_{y} \times N_{z}} \): a matrix that maps each voxel in \( D \) to the cell that contains it;
    \item \(M \in \mathbb{N}_{>0}\): an upper bound for \(N_{c}\), \JP{employed to terminate the recursive calls of Algorithm \ref{alg:cellDecomposition}};
    \JP{\item \(J \subset D\): the adjacent region of the current cell provided to the recursive calls of Algorithm \ref{alg:cellDecomposition}}; 
    \item \( p \in \{0,1\}^{4} \): a property flag vector \JP{where \(p_i\) indicates if P\(i\)} is satisfied;
    \item \(u  \in \{0,1\}^{6}\): a flag vector where each element indicates if the voxels adjacent to each boundary of \(C_{*,n}\) are uniform. The first three elements correspond to the cell boundaries in the negative \( x \), \( y \), and \( z \) directions, respectively, while the latter three correspond to the positive direction counterparts;
    \JP{\item \(F \in  \mathbb{N}_{>0}^{6\times 6}\): a set of subregions of \(D\) where \(F_{*,i}\) contains the neighboring region of the $i$-th face of \(C_{*,n}\), in the same order as in \(u\);}
    \item \(e  \in \{0,1\}^{3}\): a flag vector indicating if the current cell can expand in the positive \( x \), \( y \), and \( z \) direction, respectively;
    \item \(l  \in \mathbb{N}^{3}\): the current expansion step size for the cell in the positive \( x \), \( y \), and \( z \) direction for the current cell, respectively;
    \item \(q  \in \{0,1\}^{3}\): A flag vector \JP{where \(q_{i}\) indicates if \(l_{i}\) can be increased};
    \item \(d  \in \{1,2,3\}\): An iterator for the current expansion direction (positive \( x \), \( y \), and \( z \) direction, respectively);
    \item \(i  \in \mathbb{N}_{>0}^{3}\): The voxel indices of the current cell's starting point.
\end{itemize}
Algorithm \ref{alg:cellDecomposition}, starting \JP{ with cell index \(n = 1\) and \(C = [1,1,1,1,1,1]'\) (Lines 2-4), expands \( C_{*,n} \) in the positive \(x,y\) and \(z\) directions, \JP{one at a time}} (Lines 9-11). At each expansion step, it validates properties P1-P4 \JP{with the} following dedicated procedures, \JP{whose algorithms are outlined in Appendix \ref{append:CheckAlgorithms}}:
\begin{itemize}
    \item CheckP1(\( C_{*,n},B \)) scans the section of \(B\) contained in \( C_{*,n} \) to check for overlaps, returning a Boolean value to \(p_{1}\);
    \item CheckP2(\( C_{*,n}, D \)) scans the section of \(D\) contained in \( C_{*,n} \) to verify internal uniformity, returning a Boolean value to \(p_{2}\);
    \item CheckP3(\( C,B, D\)) identifies adjacent feasible cells from \(C_{*,n}\) by scanning the elements of \(B\) adjacent to \(C_{*,n}\), performs a mutual visibility check between the remaining adjacent cells and the $n$-th cell, and returns a Boolean value to \(p_{3}\);
    \item CheckP4(\( C_{*,n}, D \)) scans the voxels adjacent to each face of \(C_{*,n}\) to check for boundary uniformity, returning Boolean values to \(p_{4}\) and \(u\); boundaries coinciding with the grid limits are automatically considered uniform.  
\end{itemize}

If the expansion of the $n$-th cell in the $d$-th direction is successful (\GR{i.e.,} no false values in \( p \) and \(q_{d} = 1\)), \GR{the step size} \(l_{d}\) is doubled, \GR{allowing for} exponential growth of the $n$-th cell. However, if any value of \(p\) becomes 0 or a grid boundary is crossed, the last expansion is reverted to restore \JP{the $n$-th cell to \GR{its} previous valid size}\GR{. Subsequently,} \(l_{d}\) is halved and \(q_{d} = 0\) to prevent further step increases in the current direction. This \GR{sequence effectively} creates a bisection-like search for the maximum valid cell size for \(C_{*,n}\) in each direction. If \(l_{d} = 1\) when halved,  expansion in the $d$-th direction terminates (\(e_{d} = 0\)). 
\begin{footnotesize} 
    \begin{algorithm} [htbp!]
        \caption{Cell Decomposition}\label{alg:cellDecomposition}
        \begin{multicols}{2}
            \footnotesize
            \begin{algorithmic}[1]
                \Require $D \in \{0,1\}^{N_x \times N_y \times N_z}$, \(M \in \mathbb{N}_{>0}\)
                \Ensure \JP{$C \in \mathbb{N}_{>0}^{6\times N_c}$, \(o \in \{0,1\}^{N_{c}}\)} \newline
                \#Variable initialization
                \State \JP{$B \gets \mathbf{0}_{N_{x}\times N_{y} \times N_{z}}$};
                \State $n \gets 1$; $e  \gets \mathbf{1}_{1\times 3}$; \newline $p \gets \mathbf{1}_{1\times 4}$; $h \gets \mathbf{1}_{1\times 3}$; \newline \(q \gets \mathbf{1}_{1\times 3}\); \(i \gets \mathbf{1}_{1\times 3}\);
                \newline \# Main loop 
                \While{\(\exists i_1 \in \{1,...,N_x\}, i_2 \in \{1,...,N_y\}, i_3 \in \{1,...,N_z\} \mid  B_{i_1,i_2,i_3} = 0\)} 
                \newline \#Cell initialization
                \State $C_{*,n}$ $\gets$ $[i,i]^{'}$;
                \newline \#Grid boundary check and correction 
                \For{$d = 1$ to $3$}
                \If{$C_{d+3,n} \geq \mathrm{size}(D,d)$}
                \State $C_{d+3,n} \gets \mathrm{size}(D,d)$; $e_d \gets 0$;
                \EndIf
                \newline \#Cell expansion
                \If{$e_d = 1$}
                \State $C_{d+3,n} \gets C_{d+3,n} + h_d$;
                \EndIf
                \newline \#P1-P3 check and correction
                \State $p_{1} \gets \textrm{CheckP1}(C_{*,n},B)$;
                \State $p_{2} \gets \textrm{CheckP2}(C_{*,n},D)$;
                \State $p_{3} \gets \textrm{CheckP3}(C,B,D)$;
                \If{$\exists~p_{k} = 0 ~\forall~ k = 1..3$}
                \State $C_{d+3,n} \gets C(d+3,n) - l(d)$; \(q_d \gets 0\), $p \gets [1,1,1]$;
                \If{$l_d = 1$}
                    $e_d \gets 0$;
                \Else ~
                    $l_d \gets l_d/2$;
                \EndIf
                \EndIf
                \EndFor
                \newline \#P4 check and correction
                \State $p_4,\JP{u} \gets \textrm{CheckP4}(C_{*,n},D)$;
                \If{$p_{4} = 0$}
                \newline \#Cell boundary selection
                \For{\(k = 1\) to \(6\)}
                \JP{\State \(v \gets ((k-1)~\mathrm{mod}~2) + 1\);
                \State \(F_{*,k} \gets C_{*,n}\); 
                \If {\(v=k\)} 
                \State \(F_{v,k},F_{v+3,k} \gets C_{k,n} - 1\);
                \Else 
                \State \(F_{v,k},F_{v+3,k} \gets C_{k,n} + 1\);
                \EndIf
                \newline \# Recursive call to recover P4
                \If {\(u_{k} = 0\)}
                \newline \#Auxiliar vector for indexing \(D\)
                \State \(j \gets F_{*,k} \);
                \State \(J \gets D_{j_{1}:j_{4},j_{2}:j_{5},j_{3}:j_{6}}\);
                \State $t \gets \newline \textrm{CellDecomposition}(J,1)$;
                \newline \#Expansion flag update after recursive reduction
                \For{\(k = 1\) to \(3\)}
                \If{\(t_{k+3} + j_{k} \neq j_{k+3} - 1\)}
                \State \(e_{k}  \gets 0 \);
                \EndIf
                \EndFor
                \State $p_4,\JP{u} \gets \textrm{CheckP4}(C_{*,n},D)$;
                \EndIf}
                \EndFor
                \EndIf
                \newline \#End of cell expansion
                \If{$e = \mathbf{0}_{1 \times 3}$}
                \newline \# Update of cell coverage matrix \(B\)
                \For{$d = 1$ to $3$}
                \newline \#Auxiliar vector for indexing \(B\)
                \State $j_d,j_{d+3} \gets C_{d,n},C_{d+3,n}$;
                \EndFor
                \State $B_{j_1:j_4,j_2:j_5,j_3:j_6} \gets n$;
                \newline \#Increment of cell index
                \State $n \gets n + 1$;
                \newline \#Early return for recursive calls
                \If{\(n > M\)} \Return $C$ \EndIf
                \newline \#Search for starting point of the next cell within the neighborhood of current cell
                \JP{\For{\(k = 1\) to \(6\)}
                \State \(i \gets F_{1:3,k}\); 
                \If{\(B_{i_1,i_2,i_3} = 0\)} \Break \EndIf
                \EndFor}
                \newline \#Variable reset
                \State $e  \gets [1,1,1]$; $p \gets [1,1,1,1]$; $h \gets [1,1,1]$; \(q \gets [1,1,1]\);
                \EndIf
                \EndWhile
                \State \Return $C,o$;
                \end{algorithmic}
        \end{multicols}
    \end{algorithm}
\end{footnotesize} 
If \( p_{4} = 0 \), each boundary of the $n$-th cell with non-uniform adjacent voxels is individually reduced to restore P4. \JP{This is \GR{achieved} by identifying a single valid cell \(t\) within each \(F_{*,k}\) such that \(u_{k} = 0 ~\forall~ k\)}; the dimensions of \(t\) will define the new boundary. Algorithm \ref{alg:cellDecomposition} is called recursively on the subset of \(D\) delimited by \(F_{*,k}\) (named \(J\)) to find \(t\). If \( p_4 = 0 \) during the formation of \(t\), this process repeats recursively for its non-uniform boundaries, reducing a non-singleton dimension of \(J\) to 1. This recursion is guaranteed to terminate when \(J\) is reduced \JP{to have only a} single non-singleton dimension \JP{(which happens after 2 recursive calls for a 3D workspace)}, since in this case any cell in \(J\) \JP{would} only have two isolated voxels as neighbors, \GR{thereby} automatically satisfying P4. 
Once the recursive call returns the cell \(t\), the respective boundary of the parent cell \(C_{*,n}\) is reduced to match the dimensions of \(t\). The values of the expansion flag \(e\) associated with the affected coordinates of \( C_{*,n} \) are set to 0, as \( C_{*,n} \) can no longer expand in these directions without violating P3.

When all values of \( e \) are 0, the expansion of \( C_{*,n} \) is concluded\GR{. The} voxels within \( C_{*,n} \) are \GR{then} marked in the coverage array \(B\) as belonging to cell \(n\). The cell index \( n \) is \GR{subsequently} incremented, and the variables \( e, p, h, \) and \(q\) are reset to begin forming the next cell, \JP{which is initialized in one of the neighboring voxels of the current cell}. The \GR{entire} procedure finishes when all voxels in \(B\) are assigned (non-zero) or \GR{when} the maximum cell count \(M\) is exceeded \JP{(for the recursive calls)}, returning the complete cell matrix \( C \) and the \JP{obstacle cell vector \(o\)}. \JP{The dimensions of each cell in \(C\) are scaled by the pixel-to-meter ratio of the occupancy grid, \GR{ensuring} that the \GR{cell} boundaries match the real dimensions of the workspace.} 

\section{Path finding and optimization} \label{sec:pathOptimization}

After the decomposition of \( D \), the connectivity between the feasible cells in \( C \) can be represented by \JP{a line} graph \( G = \{V,E,c\} \)\GR{. The} vertex set $V \in \mathbb{I}_{1:N_{c}}$ represents the cells encoded in \(C\), and the edge set \(E \subset \{i,j \mid i,j \in V, i \neq j\} \) contains all \JP{free} cell pairs that are at least partially adjacent and whose convex hull \GR{neither contains nor intersects} any \JP{obstacle} cells. The cost function \(c: E \rightarrow \mathbb{R}_{>0} \) is defined as the traversal cost for each edge, \JP{ represented by the distance between the representative points of each connected cell \citep{Marcucci2024}}. 

From \(G\), an adjacency function \(U: V \times V \rightarrow \{0,1\} \) can be defined, where \(U(i,j) = 1\) if the $i$-th and $j$-th cell are connected in \(G\) (i.e., \(\{i,j\} \in E\)) and 0 otherwise. \GR{To} obtain a feasible polyline path, \GR{$W$}, between any pair of feasible start (\(w_{in}\)) and goal (\(w_t\)) points in \GR{the free space} \JP{\(Y_{f}\)}\GR{, let \JP{\(W = \{w_1,w_2,...,w_{N_{w}}\} \in Y_{f}^{N_{w}},\) } be} the sequence of \(N_w\) waypoints\GR{,} with \( w_1 = w_{in} \), \(w_{N_{w}} = w_t\) and \(N_{w} \geq 2\)\GR{. A path is feasible if and only if the following conditions as satisfied \citep{Ahuja1993}:}
\begin{flalign}
    & \sum_{i = 2}^{N_w}U(S_{i},S_{i-1}) = N_{w}-1,
    \label{eq:flowConstraint} \\
    &C_{1:3,S_{i}} + \epsilon \leq W_{i} \leq C_{4:6,S_{i}} - \epsilon, i = 1,2,...,N_w,
    \label{eq:boxConstraint}
\end{flalign}
with \(S \in V^{N_w}\) being the cell sequence that contains the respective waypoint in \(W\), and \(\epsilon\) being a safety margin matrix for each cell boundary. Eq. \eqref{eq:flowConstraint} ensures that \JP{\(S\) qualifies as a feasible path in \(G\)} and Eq. \eqref{eq:boxConstraint} \GR{guarantees} that each waypoint in \(W\) is contained \GR{within its corresponding} cell listed in \(S\) with safety margin \(\epsilon\). \GR{Since Property} P3 guarantees complete visibility between any two waypoints lying in adjacent cells of \(G\), verifying \eqref{eq:flowConstraint} and  \eqref{eq:boxConstraint} \JP{is} sufficient to attest \(W\) as a feasible path from \(w_{in}\) to \(w_t\), eliminating the need for explicit collision checks or \GR{the placement of} additional \JP{boundary waypoints}.


\JP{\GR{Using} Eqs. \eqref{eq:flowConstraint}-\eqref{eq:boxConstraint}, } \GR{we can formulate} \(W\) and \(S\) as the solution \GR{to} a mixed-integer optimization problem\GR{. this problem includes} \eqref{eq:flowConstraint} and \eqref{eq:boxConstraint} as feasibility constraints, with \GR{the} general formulation given by
\begin{flalign}
     & \{W^*,S^{*},N_{w},x\} = \arg\min_{W,S,N_w,x} f(W,S,x)
     \label{eq:generalMIP} \\
     & \quad~~\textrm{s.t:} ~~ g(W,S,x) \leq 0, \nonumber \\
     & \quad~~ \quad~~~ h(W,S,x) = 0, \nonumber \\
     & \quad~~ \quad~~~ W_{1} = w_{in}, W_{N_w} = w_{t}, \nonumber \\
     & \quad~~ \quad~~~\eqref{eq:flowConstraint},\eqref{eq:boxConstraint}, \nonumber
\end{flalign} 
where \(f(W,S,x)\) \GR{represents the cost} functional to be optimized by the path planning problem, \GR{while} \(x\), \(g(W,S,x)\), and \(h(W,S,x)\) represent any additional optimization variables, inequality constraints, and equality constraints of the problem\GR{, respectively.} Now let \(S^{G}\) be the shortest cell path in \(G\) obtained from graph traverse algorithms (such as Dijkstra and A*), and \GR{let} \(W^{G}\) be the shortest path constrained to \(S^G\)\GR{. If} such \GR{a} sequence is provided \textit{a priori} as a fixed optimization parameter, only \eqref{eq:boxConstraint} is required to ensure feasibility for \(W\), which reduces \eqref{eq:generalMIP} to 
\begin{flalign}
     & \{W^{G},x\} = \arg \min_{W,x} f(W,x)
     \label{eq:generalContP} \\
     & \quad~~\textrm{s.t:} ~~ g(W,x) \leq 0, \nonumber \\
     & \quad~~ \quad~~~ h(W,x) = 0, \nonumber \\
     & \quad~~ \quad~~~ W_{1} = w_{in}, W_{N_w} = w_{t}, \nonumber \\
     & \quad~~ \quad~~~\eqref{eq:boxConstraint} ,\nonumber
\end{flalign} 
by fixing the integer variables associated \JP{with} \(S\). 
\JP{Although Eq. \eqref{eq:generalMIP} can jointly optimize for the path \(W\) and the cell sequence \(S\), providing an exact solution}, its NP-hard nature makes it \JP{impractical} for large-scale problems \citep{Liberti2019}. \JP{In contrast, Eq. \eqref{eq:generalContP} \GR{offers} a more tractable approach that computes the shortest path \(W^{G}\) \JP{for a fixed cell sequence \(S^G\)\GR{. However,} \(W^{G} \equiv W^{*}\) holds only if \(S^G \equiv S^*\), a condition not guaranteed by standard} graph-based shortest path algorithms.}

\JP{To illustrate the proposed framework, the next subsection presents a path-finding problem that minimizes the Euclidean path length, along with an efficient procedure \JP{to obtain} \(P\) and \(S\).}

\subsection{\JP{Minimum-distance} path optimization}

\JP{A minimum-distance path optimization problem, derived from \eqref{eq:generalMIP} and following the formulation of \citet{Marcucci2024MISO},} can be \JP{obtained by \GR{setting} \(f(W,S,x) =\sum_{i=2}^{N_{w}}||w_i - w_{i-1}||_{2}\), with no additional variables \(x\) or constraints \(g(\cdot)\) and \(h(\cdot)\), resulting in a MISOCP. The same procedure\GR{, when} applied to the formulation based on \eqref{eq:generalContP}, \GR{yields} an SOCP. To assign meaningful traversal costs to each edge in the connectivity graph $G$ in order to obtain \(S^G\), representative points $Q_i$ are first optimized within each feasible cell to minimize the total edge distance across adjacent cells. This yields the optimization problem}
\begin{flalign}
    & Q^{*} = \arg \min_{Q} \frac{1}{2}\sum_{i=1}^{N_{C}}\sum_{j=1}^{N_{C}}U(i,j)||Q_i - Q_{j}||_{2} \label{eq:SOCPgraph} \\
     & \quad~~\textrm{s.t:} ~~C_{1:3,i} + \epsilon \leq Q_{i} \leq C_{4:6,i} - \epsilon, i \in \mathbb{I}_{1:N_{C}}. \nonumber
\end{flalign}
\JP{The optimal points $Q_i^*$ are then used to define the edge weights $c(i,j)$ as}
\begin{flalign}
    & c(i,j) = \left\{\begin{matrix}
        ||Q^*_i - Q^*_{j}||_{2}, i,j \neq \{S_1,S_{N_{W}}\}, \\
        ||w_{in} - Q^*_{j}||_{2}, i = S_{1}, \\
        ||Q^*_{i} - w_{in}||_{2}, j = S_{1},\\
        ||w_{t} - Q^*_{j}||_{2}, i = S_{N_{W}}, \\
        ||Q^*_{i} - w_{t}||_{2}, j = S_{N_{W}}, \\
    \end{matrix}\right.
    \label{eq:weightFunction}
\end{flalign}
\JP{for all pair \((i,j)\) holding \(U(i,j) = 1\) . This construction ensures that the edge weights reflect the minimum Euclidean distance between the optimized representative points of adjacent cells, while respecting the safety margin \(\epsilon\).}



\JP{The MISOCP formulation, while \GR{providing exact solution}, is intractable for large-scale problems\GR{. Conversely,} the SOCP formulation is more tractable but does not \GR{guarantee} optimality. To overcome this limitation, we propose \JP{the} KSP-SOCP approach, which \JP{integrates} Yen's k-shortest path algorithm \citep{Yen1971} with the SOCP solver. \JP{The method generates several candidate sequences} for \(S^{G}\) from Yen's algorithm, \JPL{starting at \(k = 2\) and increasing it as the \(k\)-th shortest path is found}\GR{. For each spur path evaluated, it solves} \eqref{eq:generalContP} to find the \GR{corresponding} $k$-th shortest path, and \GR{then} tests the $k$-th path found in \eqref{eq:generalContP} to improve \(W^{G}\)}. \JP{In this} way, the optimality of \(W^{G}\) is \JP{improved} while preserving the \JP{computational efficiency} of the SOCP formulation by focusing the search effort on the most promising \JP{cell sequences according to} the weight function of \(G\). \JPL{Stopping search criterias for KSP-SOCP can be defined by an upper bound on either \(k\) or in the search time, in which case the largest \(k\)  } Furthermore, \GR{since} the list of k-shortest paths in \(G\) is \JP{guaranteed to contain \(S^{*}\) for a sufficiently \JP{large} \(k\), the proposed KSP-SOCP converges to the optimal solution \(P^{*}\), thus recovering the optimality guarantees of the MISOCP formulation}. 

\section{Numerical Results} \label{sec:Results}

The cell decomposition and path planning methods were tested on 9 randomly generated city-like workspaces, each discretized \GR{into} an occupancy grid of variable size \(L\times L\times H\). For each workspace, a path from \((1,1,1)\) \GR{meters} to \((L-1,L-1,H-1)\) meters was computed using four methods: 
\JP{\begin{itemize}
    \item \JP{The} proposed KSP-SOCP using A* \citep{Hart1968} as the graph-traversing algorithm, with Euclidean distance to the goal as a heuristic\JP{,} and using MATLAB coneprog(\(\cdot\)) function to solve each SOCP;
    \item \GR{An A*-SOCP approach, which is} equivalent to KSP-SOCP with \(k = 1\);
    \item \GR{A} MISOCP solver based on the Big-M formulation, implemented in Gurobi 12.0.2 with \GR{the} MATLAB API;
    \item \GR{The} Basic Theta* \GR{planner,} \JP{employed} on the original occupancy grids, which provides a near-optimal reference in terms of path length \JP{and a time benchmark for grid-based path planning methods}.
\end{itemize} }

 All simulations were performed in MATLAB R2024b running in a single thread of an Intel i7-14700 CPU with 32 GB of RAM.

\subsection{Cell decomposition results} \label{sec:cellDecompositionResults}

The occupancy grids \JP{for all workspaces were defined with} a resolution of 1 voxel/\(\textrm{m}^{3}\), with \(L = \{100,150,...,500\}\) and \(H = 200\). \JP{The $i$-th} workspace was constructed by subdividing the XY floorpan into a \JP{\((L_{i}/50)\times(L_{i}/50)\)} grid\GR{. Each cell in this grid contains} \JP{an obstacle with a building-like shape} randomly placed \JP{inside} it. \JP{\GR{These} obstacles were generated} by stacking a random number of axis-aligned boxes \JP{with} integer dimensions, \JP{ensuring that the upper surface of each box was fully contained within that of the box below, and that the total height remained below $H$}. \JP{Fig. \ref{fig:city10x10} illustrates the $L = 500$ workspace and its corresponding feasible-cell decomposition, highlighting the overall topology of the workspaces considered and their cell decomposition according to Algorithm \ref{alg:cellDecomposition}}. Table \ref{tab:cellDecompositionResults} \JP{summarizes the} size and performance metrics \JP{obtained for} the cell decomposition of each workspace. 
\begin{table}[h!]
    \centering
    \begin{tabular}{c|c|c|c|c}
         & &  & Number & Number \\
         L & Time (s)& Memory (kB) & of cells & of edges \\
         \hline
         100 & 12 & 156 & 380 & 1149\\
         150 & 29 & 338 & 823 & 2472\\
         200 & 75 & 982 & 2385 & 7334\\
         250 & 127 & 1605 & 3888 & 12084\\
         300 & 211 & 2165 & 5249 & 16275 \\
         350 & 297 & 2686 & 6484 & 20484 \\
         400 & 399 & 3709 & 8976 & 28090 \\
         450 & 524 & 4055 & 9813 & 30719 \\
         500 & 658 & 5457 & 13208 & 41306 \\
    \end{tabular}
    \caption{Performance and size metrics of the cell decomposition procedure.}
    \label{tab:cellDecompositionResults}
\end{table}
\begin{figure*}[t!]
    \centering
    \begin{subfigure}[h!]{0.38\linewidth}
        \centering
        \includegraphics[width=\linewidth]{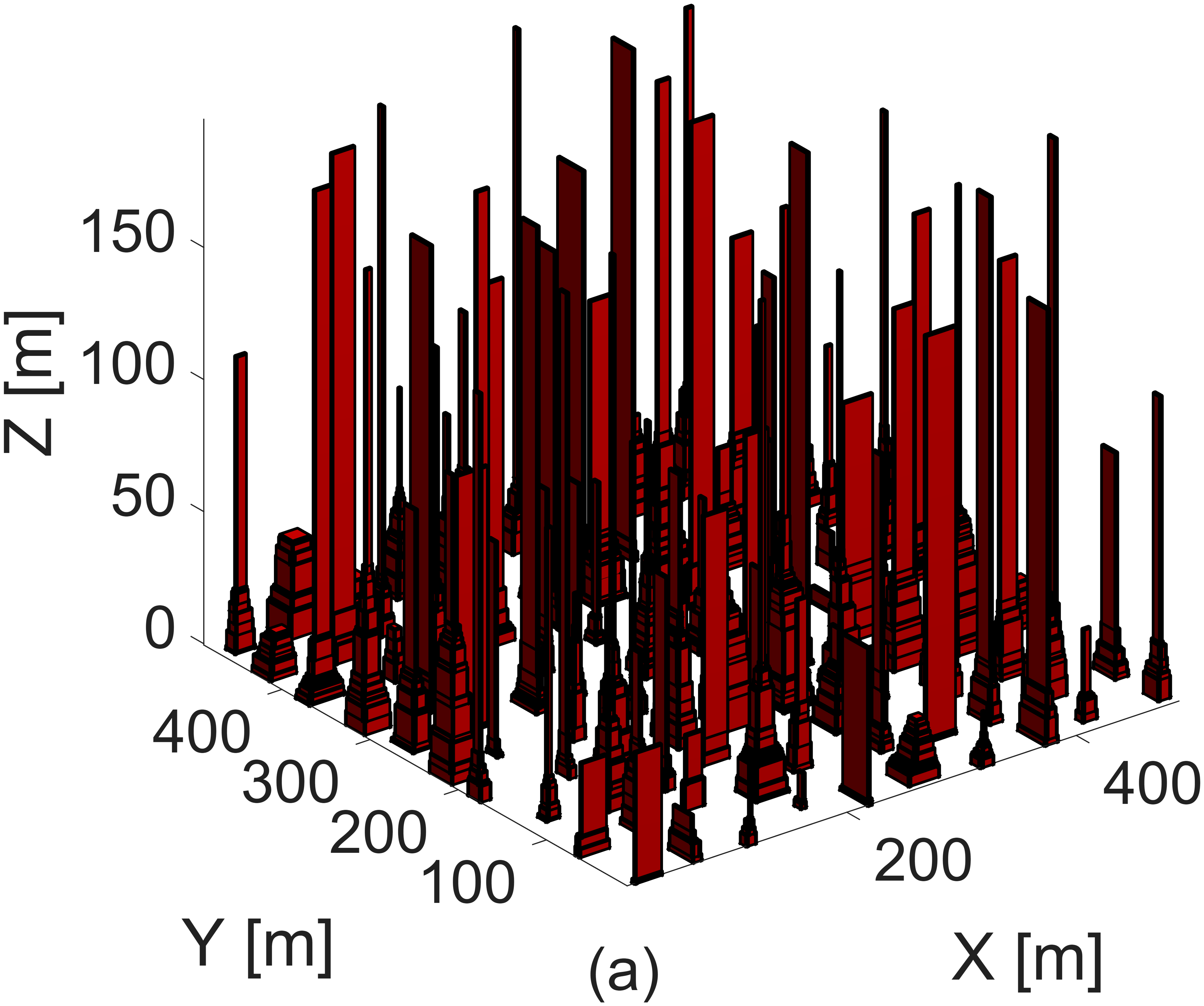}
        \label{fig:city10x10}
    \end{subfigure}
    \hspace{10mm}
    \begin{subfigure}[h!]{0.38\linewidth}
        \centering
        \includegraphics[width=\linewidth]{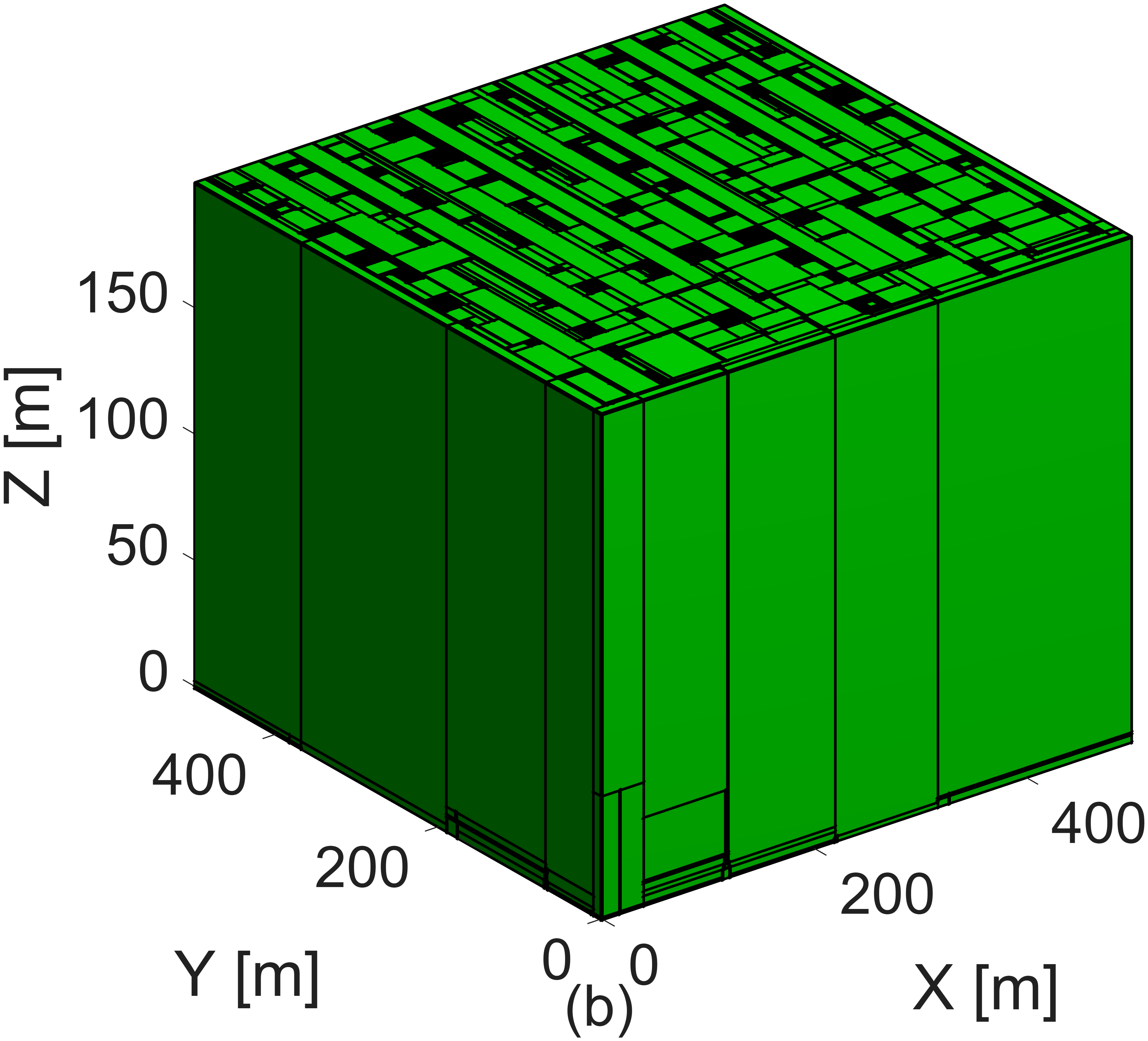}
        \label{fig:city10x10cell}
    \end{subfigure}
    \caption{\JP{(a) A 3D city-like workspace with $L = 500$. (b) Its cell decomposition with only feasible cells shown.}}
    \label{fig:city10x10}
\end{figure*}

\JP{The memory usage of the final output \GR{from} the cell decomposition procedure comprises the space occupied by the cell matrix \(C\) (\(6 \times N_c\) numbers), the occupancy vector \(o\) (\(N_{c}\) numbers), the representative points of each cell (\(3 \times N_c\) numbers), the edge list \(E\), and the weight list \(c\). All \GR{these} numbers were stored as 64-bit floating-point or integer values.}

Across all workspaces, \JP{the number of cells presented a quadratic relationship with} \(L\), fitted by \(\tilde{N_{c}}(L) = 0.0528L^2\) with \JP{coefficient of determination} \(R^{2} = 0.9958\). \JP{Since} \(H\) was fixed and the number of \JP{obstacles was} also proportional to \(L^{2}\) with a relatively homogeneous \JP{spatial} distribution, this result suggests that \(N_{c}\), \JP{in workspaces with uniform occupancy, scales with the number of voxels. For similar 3D workspaces where \(H = L\), it follows that} \(N_{C} \propto L^{3}\). 

The number of edges was nearly proportional to \(N_{c}\), with an average of 3.1 edges per cell. The mean and median number of connections \JP{per} cell were both 6, with minimum and maximum \JP{numbers} of 1 and 23, \JP{respectively}\GR{. Notably}, 99\% of \GR{the} cells \GR{had} between 1 and 19 connections.

The cell decomposition computation time, \(T_c\), also \JP{exhibited a quadratic relationship} with respect to $L$ (\( T_{c} \propto L^2\)), fitted by \(\tilde{T_{c}}(L) = 0.0025L^2\) with \(R^{2} = 0.9962\). \JP{This \(L^2\) dependence can be explained by noting that} \(T_{c}\) is \JP{essentially} given by the average computational time per cell \JP{multiplied by the number of cells}, \JP{allowing \(L^2\) to serve as a suitable proxy for \(N_c\)}. It is worth noting that the average computational time per cell may \JP{include additional contributions from} non-constant-time procedures, such as polytope intersection \JP{checks (for P3)}, 2D-3D interval scans \JP{(for P1-P2),} and recursive calls. 

\subsection{Path planning results} \label{subsec:pathPlanningResults}

\JP{In the path planning simulations, the safety margin \(\epsilon\) was \JP{set to} 1 m for every \JP{free} cell boundary adjacent to \JP{obstacle} cells, and the maximum search time was \JP{limited} to 5 minutes. \JP{To ensure consistent initialization across methods,} the MISOCP solver was warm-started \JP{using the cell-path solution obtained from A*-SOCP}}. \JP{As shown in Table \ref{tab:pathPlanningResultsThetaA*}, Basic Theta* produced paths with lengths close to} the Euclidean straight-line distance between start and goal points\GR{. However, its execution times} \JP{exhibited} \GR{a high sensitivity} to the size of \JP{the occupancy grid \(D\) and \GR{depended significantly on} whether a direct path existed or \GR{if} the algorithm needed to navigate around obstacles}.
\begin{table*}[h!]
    \centering
    \begin{tabular}{c|ccccccccc}
         & \multicolumn{2}{c}{Basic Theta*} & \multicolumn{2}{c}{A*-SOCP} & \multicolumn{3}{c}{KSP-SOCP} & \multicolumn{2}{c}{MISOCP} \\
         \hline
        $L$ & Time (s) &  Length (m) & Time (s) & Length (m) & Length (m) & \JPL{Largest \(k\)} & Memory (MB) & Length (m) & Memory (MB) \\
         \hline
         100 & 55.72 & 241.69 & 0.02 & 264.79 & 247.47 & 711  & 243.3  & 243.26 & 434.5\\
         150 & 40.72 & 288.12 & 0.08 & 321.59 & 306.57 & 145 & 189.8 & 318.24 & 1110.6\\
         200 & 2.60 & 342.95 & 0.2 & 380.07 & 375.05 & 65  & 113.0  & 380.07 & 8486.2\\
         250 & 199.03 & 402.83 & 0.3 & 495.64 & 492.52 & 43  & 148.6 & 480.47 & 21336.8\\
         300 & 84.26 & 466.18 & 0.39 & 602.43 & 555.65 & 135 & 123.9 &  \multicolumn{2}{c} {Out of memory (> 32 GB)}\\
         350 & 1275 & 531.26 & 0.59 & 697.08 & 655.27 & 19  & 133.1 &  \multicolumn{2}{c} {Out of memory (> 32 GB)}\\
         400 & 1347 & 596.67 & 0.74 & 767.52 & 762.79 & 13  & 1955.9 &  \multicolumn{2}{c}{Out of memory (> 32 GB)}\\
         450 & 589.33 & 664.39 & 1.00 & 860.00 & 848.4 & 4 & 1941.0 & \multicolumn{2}{c} {Out of memory (> 32 GB)}\\
         500 & 524.36 & 731.67 & 1.34 & 850.78 & 849.8 & 3 & 2042.7 &  \multicolumn{2}{c} {Out of memory (> 32 GB)} 
    \end{tabular}
    \caption{\JP{Performance metrics for the path planning methods evaluated. }}
    \label{tab:pathPlanningResultsThetaA*}
\end{table*}
\begin{figure}[h!]
    \centering
    \begin{subfigure}{\linewidth}
        \centering
        \includegraphics[width=\linewidth]{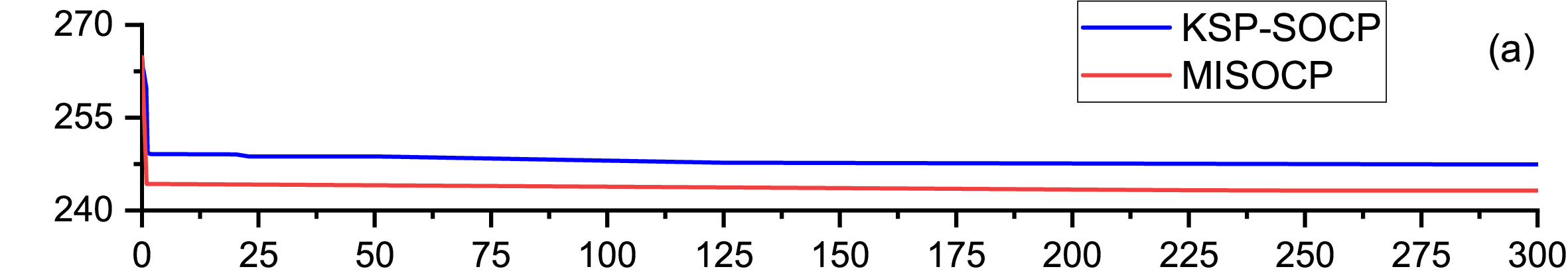}
        \vspace{-1.1em}
        \label{fig:KSPMISOCP100}
    \end{subfigure}
    \begin{subfigure}{\linewidth}
        \centering
        \includegraphics[width=\linewidth]{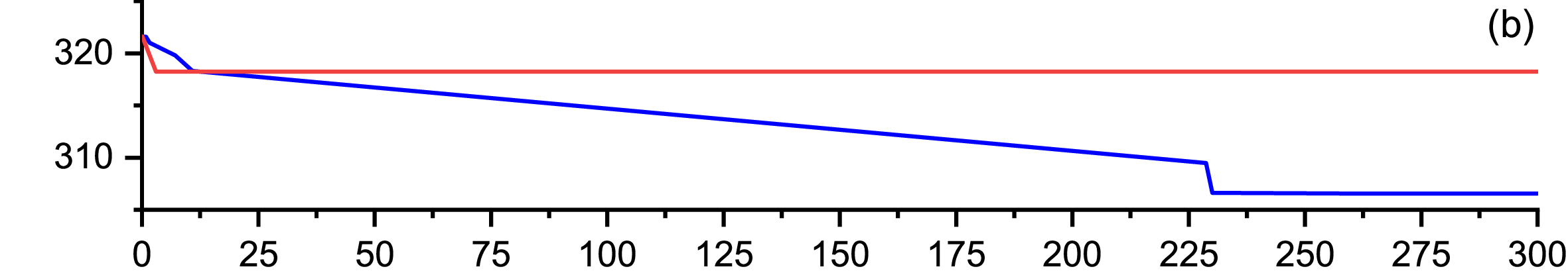}
        \vspace{-1.1em}
        \label{fig:KSPMISOCP150}
    \end{subfigure}
    \begin{subfigure}{\linewidth}
        \centering
        \includegraphics[width=\linewidth]{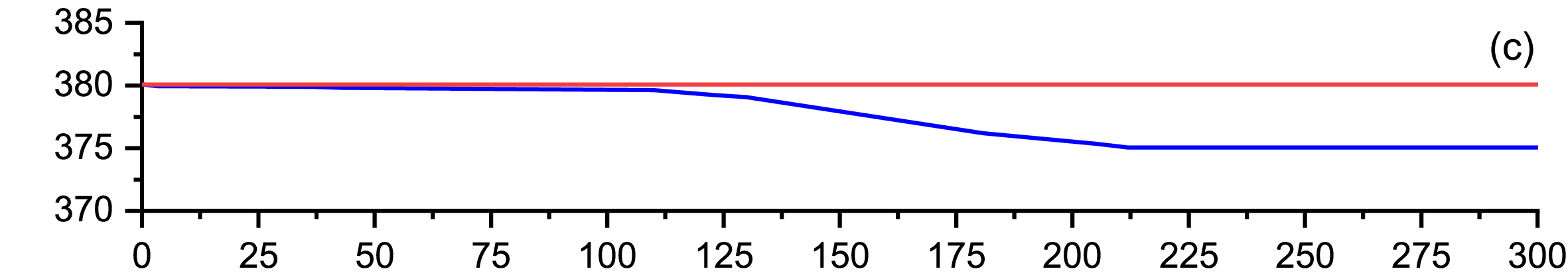}
        \label{fig:KSPMISOCP200}
        \vspace{-1.1em}
    \end{subfigure}
    \begin{subfigure}{\linewidth}
        \centering
        \includegraphics[width=\linewidth]{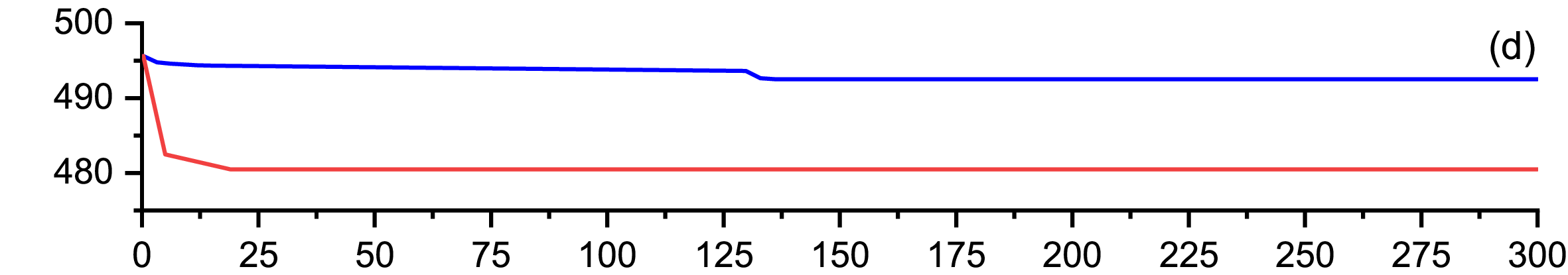}
        \label{fig:KSPMISOCP250}
        \vspace{-1.1em}
    \end{subfigure}
    \begin{subfigure}{\linewidth}
        \centering
        \includegraphics[width=\linewidth]{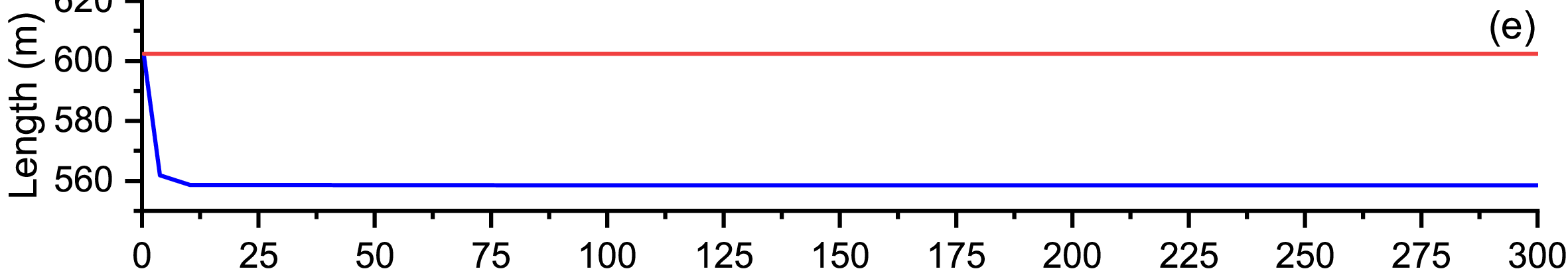}
        \label{fig:KSPMISOCP300}
        \vspace{-1.1em}
    \end{subfigure}
    \begin{subfigure}{\linewidth}
        \centering
        \includegraphics[width=\linewidth]{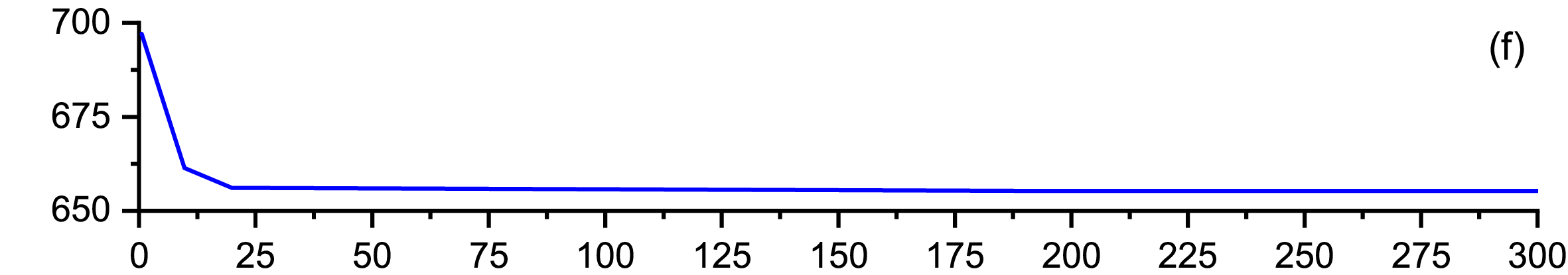}
        \label{fig:KSPMISOCP350}
        \vspace{-1.1em}
    \end{subfigure}
    \begin{subfigure}{\linewidth}
        \centering
        \includegraphics[width=\linewidth]{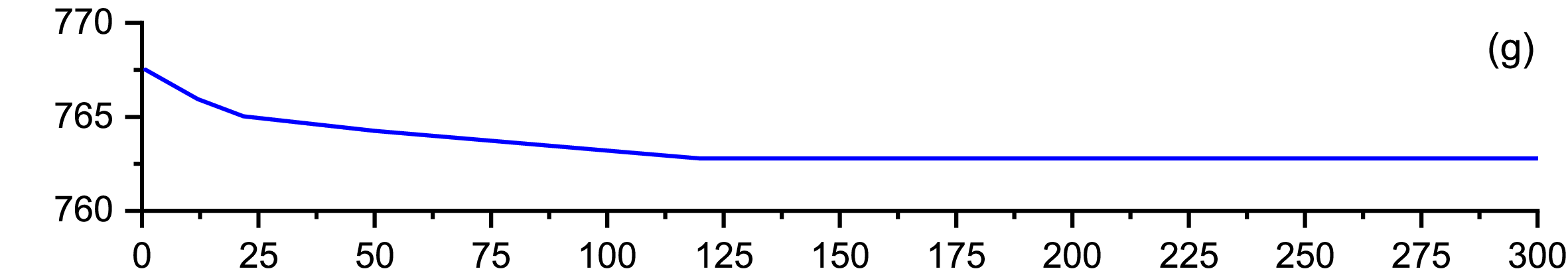}
        \label{fig:KSPMISOCP400}
        \vspace{-1.1em}
    \end{subfigure}
    \begin{subfigure}{\linewidth}
        \centering
        \includegraphics[width=\linewidth]{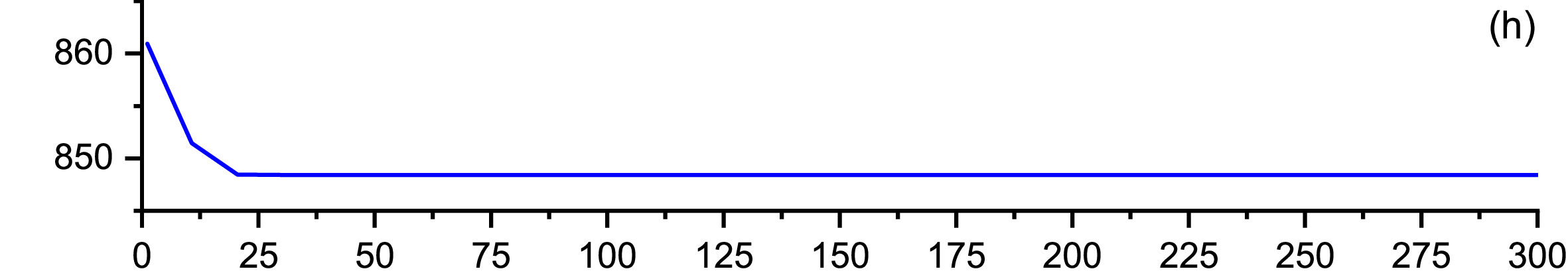}
        \label{fig:KSPMISOCP450}
        \vspace{-1.1em}
    \end{subfigure}
    \begin{subfigure}{\linewidth}
        \centering
        \includegraphics[width=\linewidth]{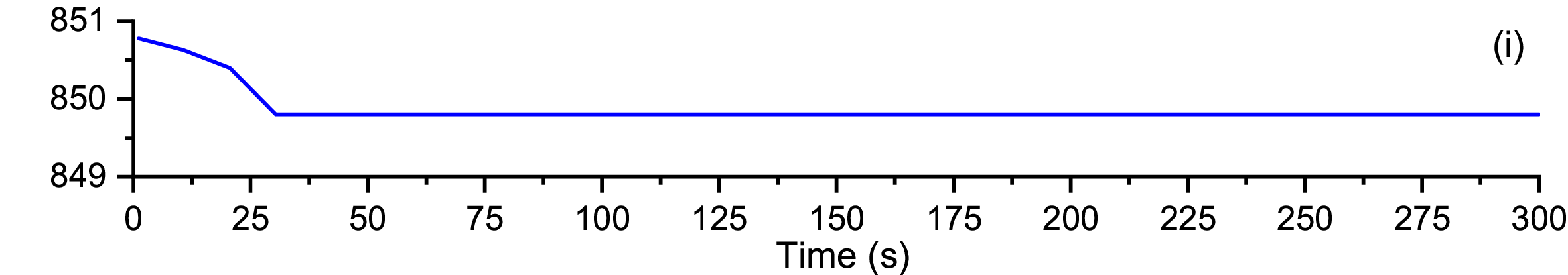}
        \label{fig:KSPMISOCP500}
        \vspace{-1.1em}
    \end{subfigure}
    \caption{\JP{Time evolution of solutions found by KSP-SOCP and MISOCP for (a) \(L = 100\), (b) \(L = 150\),..., (i) \(L = 500\)}.}
    \label{fig:timeKSPMISOCP}
\end{figure}

\GR{The} A*-SOCP \GR{approach}, on the other hand, quickly found feasible paths on \(G\), but \JP{produced} paths up to 32\% \JP{longer than those from Basic Theta*}. Both KSP-SOCP and MISOCP \JP{improved upon the A*-SOCP solutions, albeit marginally in larger workspaces}, with mixed results \JP{regarding the time evolution of the shortest path found (\GR{as} presented in Fig. \ref{fig:timeKSPMISOCP})}. In the scenarios where both methods were successfully executed, \JP{they} presented similar convergence behavior\GR{. Specifically}, MISOCP \JP{outperformed KSP-SOCP for \(L = 100\) and \(250\), \GR{while} KSP-SOCP \GR{was} ahead for \(L = 150\) and \(200\)}.\JPL{The largest value of \(k\) achieved by KSP-SOCP within the time limit showed a decreasing trend as \(L\) increased, justified by the increase with \(L\) on path length and on the time cost of running A* for each spur path.} 

The major difference \JP{appeared} in the memory usage: KSP-SOCP \JP{required} significantly less memory than MISOCP in all \JP{cases}, \JP{maintaining nearly constant memory consumption up to} $L = 400$\GR{. In sharp contrast,} \JP{MISOCP exceeded the \(32\) GB limit at \(L = 300\), \GR{which prevented} simulations at larger scales. \GR{This} difference is attributed to the expensive model building \GR{required by} MISOCP before solving and to its inherent scaling cost. Since the number of nodes, representative points, and edges in \(G\) all scale with \(L^{2}\), the memory requirements grow accordingly, quickly reaching the system's RAM limit.}

\GR{The} \JP{KSP-SOCP \GR{approach} mitigates this memory growth by embedding the flow constraints and an estimate of \GR{the} traversal distance per cell directly in \(G\), using A* to determine \(S^{G}\). This \GR{strategy} eliminates the need for integer variables and flow constraints within the SOCP, applying second-order cone constraints only to the waypoints contained in \(S^{G}\). \JPL{As a result, memory usage is limited to that required for executing A* and storing the spur and the \(k\)-shortest paths found in the allotted time limit}\GR{. Furthermore}, the smaller number of feasible paths searched by KSP-SOCP in larger workspaces partially compensates the increased memory demand due to higher \(L\)}. Therefore, the proposed KSP-SOCP approach \JP{is capable of computing feasible solutions faster than Basic Theta* and \GR{refining} the initial graph-traverse paths with a speed comparable to MISOCP, \GR{all} while using less memory\GR{. This enanles} its application to large-scale scenarios that may be unfeasible for MISOCP approaches}.

\section{Conclusions} \label{sec:Conclusions}

\JP{The extended cell decomposition algorithm efficiently partitioned dense, discretized 3D workspaces with axis-aligned box-shaped obstacles, while \GR{successfully} ensuring complete visibility and exactness properties. This \GR{achievement} enabled a simplified framework for path feasibility that \GR{was} seamlessly integrated into optimization problems. For the minimum Euclidean path distance problem, the decomposition supported the formulation of \GR{both} MISOCP and SOCP solvers for path queries in dense workspaces. The proposed KSP-SOCP method computed feasible paths as efficiently as standard SOCP-based approaches, while improving \GR{solution quality} through the convergence guarantees of MISOCP and offering superior memory efficiency for large-scale workspaces.}

\JP{In future work, we will explore alternative techniques to find suitable solutions for \(S^G\)\GR{. The aim is} to develop \GR{an} SOCP-based approach \GR{that achieves} faster convergence without compromising the memory efficiency or rapid solution generation achieved by KSP-SOCP. Furthermore, we \GR{plan to} extend the proposed cell decomposition algorithm from a discrete occupancy grid to \GR{handle} workspaces containing general polyhedral obstacles in arbitrary positions.}

\bibliography{references}             

\appendix
\section{Algorithms of CheckP1-P4 procedures}
\label{append:CheckAlgorithms}

\begin{footnotesize} 
    \begin{algorithm} [htbp!]
        \caption{CheckP1}\label{alg:CheckP1}
            \footnotesize
            \begin{algorithmic}[1]
                \Require $C \in \mathbb{N}_{>0}^{6}$, $B \in \mathbb{N}_{>0}^{N_{x}\times N_{y} \times N_{z}}$
                \Ensure $p \in \{0,1\}$
                \State \(i \gets C_{1:3}\), \(p \gets 1\)
                \While{\(i_{3} \leq C_{6}\)}
                \If{\(B_{i_{1},i_{2},i_{3}} \neq 0\)} \(p \gets 0\)
                \State \Break 
                \EndIf
                \State \(i_{1} \gets i_{1} + 1\)
                \For{\(j = 1\) to \(2\) }
                \If{\(i_{j} > C_{j+3}\)} 
                \State\(i_{j+1} \gets i_{j+1} + 1\), \(i_{j} \gets C_{j}\)
                \EndIf
                \EndFor
                \EndWhile
                \State \Return $p$
                \end{algorithmic}
    \end{algorithm}
\end{footnotesize} 
\begin{footnotesize} 
    \begin{algorithm} [htbp!]
        \caption{CheckP2}\label{alg:CheckP2}
            \footnotesize
            \begin{algorithmic}[1]
                \Require $C \in \mathbb{N}_{>0}^{6}$, $D \in \{0,1\}^{N_{x}\times N_{y} \times N_{z}}$
                \Ensure $p \in \{0,1\}$
                \State \(\mathrm{sum\_cell} \gets \mathrm{sum}(B_{C_1:C_4,C_2:C_5,C_3:C_6})\)
                \State \(\mathrm{vol\_cell} \gets \mathrm{product}(C_{4:6} - C_{1:3} + [1,1,1]')\)
                \If{\(\mathrm{sum\_cell} = 0\) \textbf{or} \(\mathrm{sum\_cell} = \mathrm{vol\_cell}\)} $p$ $\gets$ $1$
                \Else \(~p \gets 0\)
                \EndIf
                \State \Return $p$
                \end{algorithmic}
    \end{algorithm}
\end{footnotesize} 
\begin{footnotesize} 
    \begin{algorithm} [htbp!]
        \caption{CheckP3}\label{alg:CheckP3}
            \footnotesize
            \begin{algorithmic}[1]
                \Require $C \in \mathbb{N}_{>0}^{6\times n}$, $B \in \mathbb{N}^{N_{x}\times N_{y} \times N_{z}}$, $D \in \{0,1\}^{N_{x}\times N_{y} \times N_{z}}$, $n \in \mathbb{N}_{>0}$ 
                \Ensure $p \in \{0,1\}$
                \State \(adj\_c \gets \emptyset \), \(o \gets D_{C_{1,n},C_{2,n},C_{3,n}}\) 
                \For{\(k = 1\) to \(6\)}
                \State \(v \gets ((k-1)~\mathrm{mod}~2) + 1\)
                \If {\(v = k\)} \(F_{v},F_{v+3} \gets C_{k,n} - 1\)
                \Else ~\(F_{v},F_{v+3} \gets C_{k,n} + 1\)
                \EndIf
                \State \(adj\_c \gets \mathrm{append}(adj\_c,\mathrm{uniqueValues}(B_{F_{1}:F_{4},F_{2}:F_{5},F_{3}:F_{6}}))\)
                \EndFor
                \For{\(\forall ~i \in adj\_c \)}
                \State \(box \gets [\min(C_{1:3,n},C_{1:3,adj\_c_{i}})', \max(C_{4:6,n},C_{4:6,adj\_c_{i}})']'\)
                \If{\(\forall~ D_{box_{1}:box_{4},box_{2}:box_{5},box_{3}:box_{6}} = o\)} \(p \gets 1\)
                \Else
                \State \(n\_df\_vox \gets \mathrm{numElements}(D_{box_{1}:box_{4},box_{2}:box_{5},box_{3}:box_{6}} \neq o)\)
                \State \(p_x,p_y,p_z \gets \mathrm{findIndices}(D_{box_{1}:box_{4},box_{2}:box_{5},box_{3}:box_{6}} \neq o)\)
                \For{\(j = 1\) to \(n\_df\_vox\)}
                \State \(V_{*,j} \gets [p_{x_{j}} - 1,p_{y_{j}} - 1, p_{z_{j}} - 1,p_{x_{j}},p_{y_{j}},p_{z_{j}}]'\)
                \State \(A_h,b_h \gets \mathrm{convHull}(C_{*,n},C_{*,i})\)
                \State \(A_v,b_v \gets \mathrm{convHull}(V_{*,j})\)
                \State \(obstruction \gets \mathrm{checkIntersection}(A_h,b_h,A_v,b_v)\)
                \If{\(obstruction = \mathrm{True}\)}
                \Return \(p \gets 0\) \Else \(~p \gets 1\)
                \EndIf
                \EndFor
                \EndIf
                \EndFor
                \State \Return \(p \gets 1\)
                \end{algorithmic}
    \end{algorithm}
\end{footnotesize}
\begin{footnotesize} 
    \begin{algorithm} [htbp!]
        \caption{CheckP4}\label{alg:CheckP4}
            \footnotesize
            \begin{algorithmic}[1]
                \Require $C \in \mathbb{N}_{>0}^{6}$, $D \in \{0,1\}^{N_{x}\times N_{y} \times N_{z}}$
                \Ensure $p \in \{0,1\}, u \in \{0,1\}^{6}$
                \State \(p \gets 1\)
                \For{\(k = 1\) to \(6\)}
                \State \(v \gets ((k-1)~\mathrm{mod}~2) + 1\)
                \If {\(v = k\)} \(F_{v},F_{v+3} \gets C_{k} - 1\)
                \Else ~\(F_{v},F_{v+3} \gets C_{k} + 1\)
                \EndIf
                \State \(\mathrm{sum\_face} \gets \mathrm{sum}(D_{F_1:F_4,F_2:F_5,F_3:F_6})\)
                \State \(\mathrm{vol\_face} \gets \mathrm{product}(F_{4:6} - F_{1:3} + [1,1,1]')\)
                \If{\(\mathrm{sum\_face} = 0\) \textbf{or} \(\mathrm{sum\_face} = \mathrm{vol\_face}\)} $u_{k}$ $\gets$ $1$
                \Else \(~u_{k} \gets 0\), \(p \gets 0\)
                \EndIf
                \EndFor
                \State \Return \(p,u\)
                \end{algorithmic}
    \end{algorithm}
\end{footnotesize}

\end{document}